\documentclass[journal]{IEEEtran}
\usepackage{amsmath}
\usepackage{url}
\usepackage[table]{xcolor} 
\usepackage{amssymb}
\usepackage[printonlyused]{acronym}
\usepackage{subcaption}
\usepackage{csquotes}
\usepackage{dsfont}
\usepackage{amssymb}
\usepackage[flushleft]{threeparttable} 
\usepackage{booktabs}
\usepackage{graphicx}
\usepackage{booktabs, multirow}
\DeclareMathOperator*{\argmax}{argmax}

\acrodef{SNN}[SNN]{Spiking Neural Network}
\acrodefplural{SNNs}[SNNs]{Spiking Neural Networks}
\acrodef{ML}[ML]{Machine Learning}
\acrodef{ANN}[ANN]{Artificial Neural Network}
\acrodefplural{ANNs}[ANNs]{Artificial Neural Networks}
\acrodef{MLP}[MLP]{Multi Layer Perceptron}
\acrodef{CUBA-LIF}[CUBA-LIF]{Current-Based Leaky Integrate-and-Fire}
\acrodef{LIF}[LIF]{Leaky Integrate-and-Fire}
\acrodef{IF}[IF]{Integrate-and-Fire}
\acrodef{SHD}[SHD]{Spiking Heidelberg Digits}
\acrodef{N-MNIST}[N-MNIST]{Neuromorphic MNIST}
\acrodef{FSNN}[FSNN]{Feed-forward SNN}
\acrodef{RSNN}[RSNN]{Recurrently-connected SNN}
\acrodef{RCNN}[RCNN]{Recurrently Connected Neural Networks}
\acrodef{BPTT}[BPTT]{Backpropagation Trought Time}
\acrodef{AI}[AI]{Artificial Intelligence}

\begin{document}

\title{Impact of spiking neurons leakages and \\ network recurrences on event-based \\ spatio-temporal pattern recognition}

\author{
\IEEEauthorblockN{Mohamed Sadek Bouanane$^1$, Dalila Cherifi$^1$, Elisabetta Chicca$^{2,3}$, Lyes Khacef$^{2,3,*}$}

\IEEEauthorblockA{
\textit{$^1$Institute of Electrical and Electronic Engineering, Univesity of Boumerdes, Algeria.} \\
\textit{$^2$Bio-Inspired Circuits and Systems (BICS) Lab. Zernike Institute for Advanced Materials, \\ University of Groningen, the Netherlands.} \\
\textit{$^3$Groningen Cognitive Systems and Materials Center (CogniGron), University of Groningen, the Netherlands.} \\
$^*$l.khacef@rug.nl}
}


\maketitle

\begin{abstract}
Spiking neural networks coupled with neuromorphic hardware and event-based sensors are getting increased interest for low-latency and low-power inference at the edge. 
However, multiple spiking neuron models have been proposed in the literature with different levels of biological plausibility and different computational features and complexities. Consequently, there is a need to define the right level of abstraction from biology in order to get the best performance in accurate, efficient and fast inference in neuromorphic hardware. 
In this context, we explore the impact of synaptic and membrane leakages in spiking neurons. We confront three neural models with different computational complexities using feedforward and recurrent topologies for event-based visual and auditory pattern recognition. 
Our results show that, in terms of accuracy, leakages are important when there are both temporal information in the data and explicit recurrence in the network. In addition, leakages do not necessarily increase the sparsity of spikes flowing in the network. We also investigate the impact of heterogeneity in the time constant of leakages, and the results show a slight improvement in accuracy when using data with a rich temporal structure. 
These results advance our understanding of the computational role of the neural leakages and network recurrences, and provide valuable insights for the design of compact and energy-efficient neuromorphic hardware for embedded systems.
\end{abstract}


\IEEEpeerreviewmaketitle


\section{Introduction}
\IEEEPARstart{O}{ver} the last decade, \acp{ANN} have been increasingly attracting interest in both academia and industry as a consequence of the explosion of open data and the high computing power of today’s computers for training and inference. 
The state-of-the-art performance of deep neural networks on various pattern recognition tasks has given neural networks the leading role in \ac{ML} algorithms and \ac{AI} research.
However, the technological drive that has supported Moore’s Law for fifty years and the increasing computing power of conventional processors is reaching a physical limit and is predicted to flatten by 2025 \cite{Shalf20}. 
Hence, deep learning progress with current models and implementations will become technically, economically and environmentally unsustainable \cite{Thompson_etal20,Thompson_etal21}.
This limit is particularly prohibitive when targeting edge application in embedded systems with severe constraints in latency and energy consumption \cite{Rabaey_etal19}.

Neuromorphic computing is a promising solution that takes inspiration from the biological brain which can reliably learn and process complex cognitive tasks at a very low power consumption. 
On the one hand, neuromorphic sensors are event-based sensors and capture information with a high spatio-temporal sparsity and high temporal resolution at low-latency and low-power consumption \cite{Liu_etal10,Gallego_etal22}.
On the other hand, neuromorphic processors are asynchronous and use parallel and distributed implementations of synapses and neurons where memory and computation are co-localized \cite{Mead_Conway80,Chicca_etal14b}, hence adapting the hardware to the computation model \cite{Schuman2017,Bouvier_etal19}.
\acp{SNN} are the the generation of artificial neural models \cite{Maass97} that are investigated to exploit the advantages of event-based sensing and asynchronous processing at the algorithmic level.

Inspired from the neuroscience literature, \acp{SNNs} show promising performance in embedded spatio-temporal pattern recognition \cite{Davies_etal21}. 
For example, compared to a conventional approach using formal neural networks on an embedded Nvidia Jetson GPU, \acp{SNN} on the Intel Loihi neuromorphic chip \cite{Davies2018} achieve a gain in energy-efficiency of $~30 \times$ in multimodal (vision and EMG) hand gesture recognition \cite{Ceolini_etal20} and $~500 \times$ in tactile braille letters recognition \cite{Muller-Cleve2022}, at the cost of a loss in accuracy depending on the application. 
Multiple models of spiking neurons have been proposed in the literature \cite{Hodgkin_Huxley52,Kistler_etal97,Izhikevich03} and implemented in hardware \cite{Indiveri_etal11} with different levels of biological plausibility and computational complexity.
However, there is a lack of understanding of how each of the factors determining the biological neuronal response can be effectively used in learning and inference. 
A key question for advancing the field is therefore to identify the right level of abstraction inspired from biology to achieve the best inference performance within strict constrains in speed/latency and power efficiency on neuromorphic hardware.

This work attempts to partially answer this question by studying the effect of spiking neurons leakages in feedforward and recurrent neural networks for event-based visual and auditory pattern recognition tasks, in terms of accuracy and spiking activity.
Today, digital neuromorphic chips from academia and industry use both non-leaky (e.g. SPLEAT \cite{Abderrahmane_etal22} and DynapCNN \cite{Liu_etal19}) and leaky (e.g. MorphIC \cite{Frenkel_etal19} and Loihi \cite{Davies2018}) spiking neurons.
Understanding the computational role of the leakages provides insights for the hardware architecture of neuromorphic processors as they require extra circuitry overheads \cite{Khacef_Abderrahmane2018}.
In Section 2 we introduce the spiking neuron models and present the training methodology. Then, in section 3 we present experiments on the spiking neuron leakages, the network typologies and the time constants heterogeneity, and provide a detailed analysis of the obtained results. Finally, in sections 4 and 5 we discuss the results, highlighting the main insights, limits and outlook of our work.


\section{Methods}
In this section, we introduce the used spiking neuron models that are characterized with different levels of biological abstraction. We also introduce the surrogate gradient decent approach used in this work to overcome the all-or-nothing behavior of the binary spiking non-linearity.

\subsection{Spiking neuron models}
The standard spiking neuron model is formally described as a time continuous dynamical system with the differential equation \cite{Gerstner_Kistler_Naud_Paninski2014}:

\begin{equation}
    \tau_{mem}\frac{\mathrm{d} U_{i}^{(l)}(t)}{\mathrm{d} t} = -(U_{i}^{(l)}(t)-U_{rest})+RI_{i}^{(l)}(t)
    \label{mem-po}
\end{equation}

\noindent where $U_{i}(t)$ is the membrane potential that characterizes the hidden state of the neuron, $U_{rest}$ is the resting potential, $\tau_{mem}$ is the membrane time constant, $R$ is the input resistance, and $I_{i}(t)$ is the input current. The hidden state of each neuron, however, is not directly communicated to downstream neurons. When the membrane potential $U_{i}$ reaches the firing threshold $\vartheta$, the neuron fires an action potential (or a \enquote{spike}) and the membrane potential $U_{i}$ is reset to its resting potential $U_{rest}$. 
If we consider spikes to be point processes for which their spike width is zero in the limit, then a spike train $S_{j}^{(l)}(t)$ is denoted with the sum of Dirac delta functions $S_{j}^{(l)}(t)=\sum_{s\in C_{j}^{(l)}} \delta (t-s)$ such that $s$ iterate over the firing times $C_{j}^{(l)}$ of neuron $j$ from layer $l$.
Spikes are communicated to downstream neurons and trigger postsynaptic currents.
A common first-order approximation to model the temporal dynamics of postsynaptic currents are exponentially decaying currents that sum linearly:

\begin{equation}
    \frac{\mathrm{d} I_{i}(t)^{(l)}}{\mathrm{d} t} = -\frac{I_{i}^{(l)}(t)}{\tau _{syn}} + \sum_{j}W_{ij}^{(l)}S_{j}^{(l-1)}(t) + \sum_{j}V_{ij}^{(l)}S_{j}^{(l)}(t)
    \label{syn}
\end{equation}

\noindent where we have introduced the synaptic decay time constant $\tau_{syn}$, and the synaptic weight matrices: $W_{ij}^{(l)}$ for feed-forward connections, and $V_{ij}^{(l)}$ for explicit recurrent connections within each layer. 

\par It is customary to approximate the solutions to the above equations in discrete time assuming a small simulation time step $\Delta t > 0$. With no loss of generality, we assume $U_{rest}=0$, $R=1$, and the firing threshold $\vartheta=1$. The output spike train $S_{i}^{(l)}[t]$ of neuron $i$ in layer $l$ is expressed as $S_{i}^{(l)}[t] \equiv \Theta (U_{i}^{(l)}[t] - \vartheta)$ where $\Theta$ is the Heaviside step function such that $S_{i}^{(l)}[t] \in\left \{ 0,1 \right \}$. $t$ is used to denote the time step to indicate discrete time. The synaptic and membrane dynamics expressed respectively by Equation \ref{syn} and Equation \ref{mem-po} become \cite{Neftci_Mostafa_Zenke2019}:

\begin{equation}
    I_{i}^{(l)}[t] = \alpha I_{i}^{(l)}[t-1] + \sum_{j}W_{ij}^{(l)}S_{j}^{(l-1)}[t-1]+ \sum_{j}V_{ij}^{(l)}S_{j}^{(l)}[t-1]
    \label{syn-dic}
\end{equation}

\begin{equation}
    U_{i}^{(l)}[t] = (\beta U_{i}^{(l)}[t-1] + I_{i}^{(l)}[t]) \times (1-S_{i}^{(l)}[t-1]) 
    \label{mem-disc}
\end{equation}

\noindent where the decay strengths are given by $\alpha \equiv  e^{-\frac{\Delta t}{\tau _{syn}}}$ and $\beta \equiv  e^{-\frac{\Delta t}{\tau _{mem}}}$ , such that $0<\alpha<1$ and $0<\beta<1$ for finite and positive $\tau_{syn}$ and $\tau_{mem}$.

There exists many extensions and variations of spiking neurons models. In order to find the right level of abstraction from biology and get the best performance in accurate, efficient and fast inference, we will derive and confront three variations with variable degrees of biological plausibility: the \ac{CUBA-LIF}, the \ac{LIF}, and the \ac{IF}.  

\subsubsection{Current-Based Leaky Integrate-and-Fire (CUBA-LIF)}
The \ac{CUBA-LIF} neuron is the most biologically plausible model among the three models considered in this work. It accounts for the temporal dynamics of the postsynaptic current. This neuron model is governed by Equations \ref{cuba-syn-dic} and \ref{cuba-mem-disc}. It has two exponentially decaying terms: $\alpha I_{i}$ and $\beta U_{i}$. The degree of the exponential decay of $I_{i}$ and $U_{i}$ is determined by the synaptic time constant $\tau _{syn}$ and membrane time constant $\tau_{mem}$, respectively. Figure \ref{fig:CUBALIF} illustrate the dynamics of a \ac{CUBA-LIF} neuron for some random input stimuli.

\begin{equation}
    I_{i}^{(l)}[t] = \alpha I_{i}^{(l)}[t-1] + \sum_{j}W_{ij}^{(l)}S_{j}^{(l-1)}[t-1]+ \sum_{j}V_{ij}^{(l)}S_{j}^{(l)}[t-1]
    \label{cuba-syn-dic}
\end{equation}

\begin{equation}
    U_{i}^{(l)}[t] = (\beta U_{i}^{(l)}[t-1] + I_{i}^{(l)}[t]) \times (1-S_{i}^{(l)}[t-1]) 
    \label{cuba-mem-disc}
\end{equation}

\subsubsection{Leaky Integrate-and-Fire (LIF)}
The \ac{LIF} neuron model is a simplification of the \ac{CUBA-LIF} and it is widely used in computational neuroscience to emulate the dynamics of biological neurons \cite{Izhikevich2004}. It integrates the input over time with a leakage such that the internal state represented by the membrane potential goes down exponentially. As shown in Figure \ref{fig:LIF}, subsequent input spikes must be maintained for the state not to go to zero. In discrete time, the dynamics the \ac{LIF} neuron are governed by Equations \ref{lif-syn-disc} and \ref{lif-mem-disc}:

\begin{equation}
    I_{i}^{(l)}[t] = \sum_{j}W_{ij}^{(l)}S_{j}^{(l-1)}[t-1] + \sum_{j}V_{ij}^{(l)}S_{j}^{(l)}[t-1]
    \label{lif-syn-disc}
\end{equation}

\begin{equation}
    U_{i}^{(l)}[t] = (\beta U_{i}^{(l)}[t-1] + I_{i}^{(l)}[t]) \times (1-S_{i}^{(l)}[t-1]) 
    \label{lif-mem-disc}
\end{equation}

\subsubsection{Integrate-and-Fire (IF)}
The \ac{IF} neuron is a further simplification and the least biologically plausible model considered in this work. The \ac{IF} model can be concisely described as a \ac{LIF} neuron with no leak. It behaves as a standard integrator that keeps a running sum of its input. Thus, the internal state of the neuron is the mathematical integral of the input \cite{Eliasmith2013}. \ac{IF} neurons do not have any inherent temporal dynamics. In discrete time, \ac{IF} dynamics are governed by Equations \ref{if-syn-disc} and \ref{if-mem-disc}:

\begin{equation}
    I_{i}^{(l)}[t] = \sum_{j}W_{ij}^{(l)}S_{j}^{(l-1)}[t-1] + \sum_{j}V_{ij}^{(l)}S_{j}^{(l)}[t-1]
    \label{if-syn-disc}
\end{equation}

\begin{equation}
    U_{i}^{(l)}[t] = (U_{i}^{(l)}[t-1] + I_{i}^{(l)}[t]) \times (1-S_{i}^{(l)}[t-1]) 
    \label{if-mem-disc}
\end{equation}

\begin{figure*}[h]
\centering
\begin{subfigure}{0.3\textwidth}
  \centering
  \includegraphics[width=\textwidth]{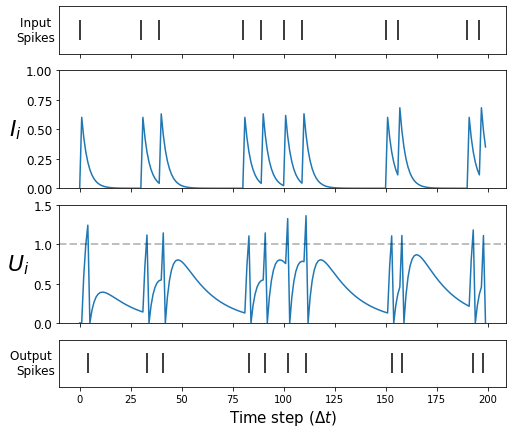}
  \caption{CUBA-LIF}
  \label{fig:CUBALIF}
\end{subfigure}
\hspace{0.5em}
\begin{subfigure}{0.3\textwidth}
  \centering
  \includegraphics[width=\textwidth]{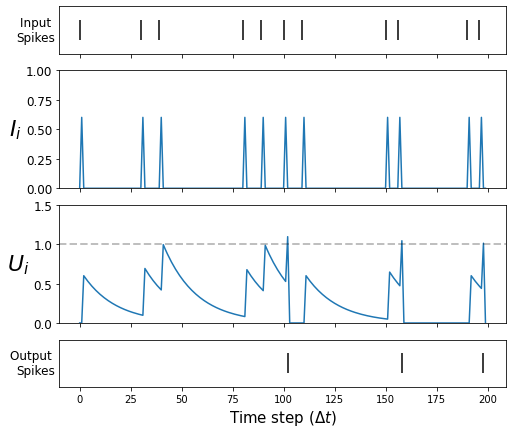}
  \caption{LIF}
  \label{fig:LIF}
\end{subfigure}
\hspace{0.5em}
\begin{subfigure}{0.3\textwidth}
  \centering
  \includegraphics[width=\textwidth]{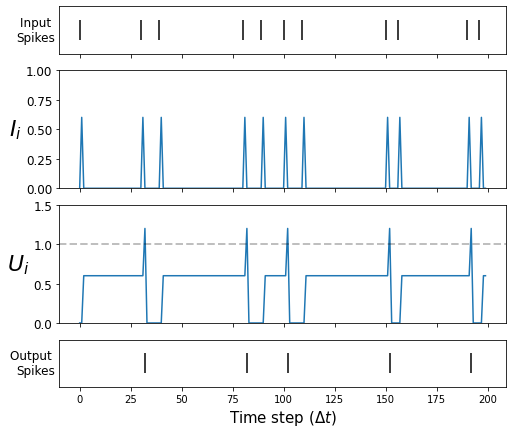}
  \caption{IF}
  \label{fig:IF}
\end{subfigure}
\caption{Synaptic current and membrane potential dynamics of each spiking neuron model in response to the same input spikes. Only when the membrane potential reaches the neuronal firing threshold (dashed line), output spikes are generated}
\label{fig:neuron-dynamics}
\end{figure*}

Equations \ref{if-syn-disc} and \ref{if-mem-disc} do not have the decay (i.e., leak) parameters $\alpha$ and $\beta$. The parameter $\alpha$ is set to $zero$ which causes the synapse to have an infinite leak. The current pulse width is short, it effectively looks like a weighted spike. $\beta$ on the other hand is set to $one$, which causes the membrane potential to remain constant between two consecutive spikes. Figure \ref{fig:IF} illustrates the dynamics of an \ac{IF} neuron for some random input stimuli.

\subsection{Supervised learning in \acp{SNNs}} 
The choice of the three spiking neurons models considered is motivated by the intention of mapping existing machine learning methods to train \acp{SNNs}. The aim of learning is to minimize a loss function $\mathcal{L}$ over the entire dataset. The gradient-based method, namely \ac{BPTT} \cite{Goodfellow2016} was used. Before \ac{BPTT} can be applied to \acp{SNNs}, however, a serious challenge regarding the non-differentiability of the spiking non-linearity needs to be overcome.

\ac{BPTT} requires the calculation of the derivative of the neural activation function. For a spiking neuron, however, the derivative of $S[t] =  \Theta (U[t] - \vartheta)$ is zero everywhere except at $U=\vartheta$, where it tends to infinity as shown in Equation \ref{bp-0-inf}. This means the gradient will almost always be zero and no learning can take place. This behaviour of the binary spiking non-linearity makes \acp{SNNs} unsuitable for gradient based optimization and it is known as the \enquote{\textit{dead neuron problem}}.

\begin{equation}
    \frac{\partial \mathcal{L}}{\partial W} = \frac{\partial \mathcal{L}}{\partial S}  \underbrace{\frac{\partial S}{\partial U}}_\text{\{0,$\infty$\}}  \frac{\partial U}{\partial I}  \frac{\partial I}{\partial W}
    \label{bp-0-inf}
\end{equation}

In this work, we used a surrogate gradient approach \cite{Neftci_Mostafa_Zenke2019} to provide a continuous relaxation to the real gradients. In other words, we keep the Heaviside step function the way it is during the forward pass and change the derivative term ${\partial S}/{\partial U}$ with something that does not stop learning during the backward pass. Specifically, we selected the fast sigmoid function $\tilde{S}$ to smooth out the gradient of the Heaviside function: 

\begin{equation}
    \tilde{S} = \sigma(U_{i}^{(l)})=\frac{U_{i}^{(l)}}{1+\tilde{\beta} \left | U_{i}^{(l)} \right |}
\end{equation}

\noindent where $\tilde{\beta}$ is the steepness parameter that modulates how smooth the surrogate function is.

In this work, cross entropy \textit{max-over-time} loss function \cite{Cramer2022} is chosen. When called, the maximum membrane potential value for each output neuron in the readout layer is sampled and passed through the loss function. This cross entropy loss encourages the maximum membrane potential of the correct class to increase, and suppresses the maximum membrane potential of incorrect classes. On data with batch size of $N_{batch}$ and $N_{class}$ output classes, $\{(x_{s},y_{s}) \mid s=1,...,N_{batch} ;  y_{s} \in \{1,...,N_{class}\}\}$ the loss function takes the form:

\begin{equation}
    \mathcal{L} = -\frac{1}{N_{batch}}\sum_{s=1}^{N_{batch}} \mathds{1}(i=y_{s}) \log \left \{ \frac{\exp(U_{i}^{(L)}[\tilde{t_{i}}])}{\sum_{i=0}^{N_{class}}\exp(U_{i}^{(L)}[\tilde{t_{i}}])} \right \}
    \label{cross-entropy}
\end{equation}

\noindent where $\mathds{1}$ is the indicator function, and $\tilde{t}$ is the time step with the maximum membrane potential for each readout unit in the readout layer $L$, such that $\tilde{t_{i}} = \argmax_{t} U_{i}^{(L)}[t]$. \\
The cross entropy in Equation \ref{cross-entropy} is minimized using the Adamax optimizer \cite{Kingma_Diederik_Ba2014}. 


\section{Experiments and results}
This section present all the experiments we conducted in order to understand the effect of spiking neurons leakages and network recurrences for spike-based spatio-temporal pattern recognition and gives a detailed analysis of the results we obtained. 

\subsection{Experimental setup}
We investigated the role of neurons leakages, network recurrences and neural heterogeneity by training \acp{SNNs} to classify visual and auditory stimuli with varying degrees of temporal structure. We adopted a necessary and sufficient minimal architecture widely used as an universal approximator \cite{Maass_Wolfgang_Natschlager2002,Neftci_Mostafa_Zenke2019} consisting of three layers of spiking neurons: an input layer, a hidden layer with or without recurrent connections, and a readout layer used to generate predictions. This is a universal approximator, where the readout layer consists of neurons that do not spike. Two training approaches were applied: standard training only modifies the synaptic weights, while heterogeneous training affects both the synaptic weights and the time constants. 

We used two datasets. \ac{N-MNIST} \cite{Orchard2015} contains mostly spatial information. It features visual stimuli and has minimal temporal structure, as its samples are generated from static images by moving a neuromorphic vision sensor over each original MNIST sample. Therefore, the spike rate of the input neurons has sufficient information about the pattern, while the temporal component is strictly related to the movements of the vision sensor. By contrast, the \ac{SHD} \cite{Cramer2022} is auditory and has a rich temporal structure. It was generated using an artificial cochlea, where the spike timing of the input neurons is necessary to recognize each pattern \cite{Perez-Nieves_2021}.

To allow for fair baselines comparison of performance in accuracy with previous works, we used the same train/test split suggested by the corresponding dataset authors in all of our experiments. The network architectures and common hyper-parameters in our experiments such as the batch size, number of epochs, learning rate $\eta$, and steepness parameter $\tilde{\beta}$ were tuned according to state-of-the-art results obtained from the literature \cite{Cramer2022, Perez-Nieves_2021, Chowdhury_Lee2021}, as well as our own preliminary experiments. Table \ref{tab:parameters} gives a summary of all parameters used for our experiments. The performance of each configuration is quantified in terms of testing accuracy and sparsity as an estimation for dynamic energy-efficiency in neuromorphic hardware. We note that all reported error measures in this work correspond to the standard deviation of three experiments with different random initialization for the trained parameters.

\begin{table}[h]
\centering
\caption{Hyperparameters used in our experiments.}
\begin{tabular}{lcc} 
\hline
                     & \textbf{\ac{N-MNIST}}      & \textbf{\ac{SHD}}          \\ 
\hline
\textbf{Train/Test split}     & 60,000/10,000 & 8332/2088   \\
\textbf{Network architecture} & 2312-200-10 & 700-200-20  \\
\textbf{Learning rate ($\eta$)}~       & 5$\times10^{-3}$        & 2$\times10^{-4}$        \\
\textbf{Time step ($\Delta t$)}            & 14ms         & 14ms         \\
\textbf{Steepness parameter ($\tilde{\beta}$)}  & 100         & 100         \\
\textbf{Batch size }          & 256         & 128         \\
\textbf{Epochs}               & 50          & 200         \\
\hline
\end{tabular}
\label{tab:parameters}
\end{table}

\subsection{Impact of the neural leakages}
To assess the effects of the membrane and synaptic leakages, we started by confronting the three concerned neuron models: \ac{CUBA-LIF}, \ac{LIF}, and \ac{IF} in a \ac{FSNN} using both datasets. Only synaptic weights are learned and leakage parameters are treated as hyper-parameters and chosen to be homogeneous (i.e. the same for all neurons). Leakage parameters $\alpha$ and $\beta$ are tuned using the synaptic time constant $\tau_{syn}$ and the membrane time constant $\tau_{mem}$, respectively, as described by Equations \ref{syn-dic} and \ref{mem-disc} in the previous chapter. 

\subsubsection{Accuracy analysis in \ac{FSNN}}
We started by the \ac{CUBA-LIF} neuron where both leakages are of concern. This model can have a wide range of $\tau_{syn}$ and $\tau_{mem}$. We performed a grid search across a number of time constants by fixing one and changing the other. Grid search is a simple hyper-parameters tuning technique that helped us evaluate the model for a wide range of combinations to get a good understanding of the slope of change in accuracy.

Table \ref{sub:shd-cuba-fsnn} shows the \ac{SHD} testing accuracy results for the chosen different combinations of $\tau_{mem}$ and $\tau_{syn}$. We can see from the time constants sweeps that $\tau_{mem}$ values below $420ms$ result is a significant decrease in accuracy. It is also clear that the best results seem to push $\tau_{syn}$, and hence $\alpha$, close to zero with $\tau_{mem} \geq 420ms$. In other words, \ac{CUBA-LIF} performs better when its dynamics are close to those of the \ac{LIF} neuron. This trend is also observed for the \ac{N-MNIST} as shown in Table \ref{sub:mnist-cuba-fsnn}. However, we can see a 31.55\% drop between the best accuracy that reached 76.94\% $\pm$ 1.13\% and the 45.39\% $\pm$ 1.64\% worst accuracy for \ac{SHD}, while only a 1.59\% difference between the 97.41\% $\pm$ 0.07\% best and the 95.82\% $\pm$ 0.11\% worst accuracy for \ac{N-MNIST}. This suggests that the leakages seem to have greater impact on data with rich temporal structure, than on data that is intrinsically spatial and low in temporal structure.

\par The \ac{LIF} neuron has an infinite synaptic leak with $\tau_{syn}=0$ and a tunable membrane leak. So we varied $\tau_{mem}$ as a hyperparameter across a range of values. The results of our experiments presented in Table \ref{sub:shd-cuba-fsnn} show that the \ac{LIF} neuron achieved higher accuracy than its \ac{CUBA-LIF} counterpart for both datasets. but it also resulted in the lowest accuracies for very small values of $\tau_{mem}$.
\par For the \ac{IF} case, there is an infinite synaptic leak similar to that of the \ac{LIF} and no membrane leak. so the only possible values for time constants are $\tau_{mem}=\infty$ and $\tau_{syn}=0$ which corresponds to $\beta=1$ and $\alpha=0$ respectively. In spite of its simplicity and lack of temporal dynamics, \ac{IF} neuron was able to match or even outperform the other models by reaching a testing accuracy of 78.36\% $\pm$ 0.87\% for \ac{SHD}  and 97.50\% $\pm$ 0.06\% for \ac{N-MNIST} as shown in Table \ref{tab:cuba-fsnn}. This result suggests that introducing inherent temporal dynamics and increasing neuronal complexity does not necessarily lead to an improved classification accuracy even for data with rich temporal structure when using a feed-forward network.

\begin{table*}[!htp]\centering
\caption{Three neuron models accuracy in FSNN.}\label{tab:cuba-fsnn}
\scriptsize
\begin{subtable}{\textwidth}
\centering
\caption{SHD}
\label{sub:shd-cuba-fsnn}
\begin{threeparttable}
\begin{tabular}{lc|ccccc}\toprule
\textbf{} &\textbf{LIF} &\multicolumn{4}{c}{\textbf{CUBA-LIF}} \\
\textbf{(ms)} &\textbf{$\tau_{syn}=0$ $(\alpha \approx 0)$} &\textbf{$\tau_{syn}=14$ $(\alpha \approx 0.368)$} &\textbf{$\tau_{syn}=28$ $(\alpha \approx 0.606)$} &\textbf{$\tau_{syn}=70$ $(\alpha \approx 0.818)$} &\textbf{$\tau_{syn}=140$ $(\alpha \approx 0.905)$} \\\midrule
\textbf{$\tau_{mem}=14$ $(\beta \approx 0.368)$} &\cellcolor[HTML]{ff0000}38.24\% &\cellcolor[HTML]{ff2d00}45.39\% &\cellcolor[HTML]{ff4600}49.39\% &\cellcolor[HTML]{ff7100}56.14\% &\cellcolor[HTML]{ff8b00}60.19\% \\
\textbf{$\tau_{mem}=70$ $(\beta \approx 0.818)$} &\cellcolor[HTML]{ff5d00}52.88\% &\cellcolor[HTML]{ff6100}53.60\% &\cellcolor[HTML]{ff8b00}60.15\% &\cellcolor[HTML]{ff9400}61.63\% &\cellcolor[HTML]{ffa500}64.24\% \\
\textbf{$\tau_{mem}=140$ $(\beta \approx 0.905)$} &\cellcolor[HTML]{ffae00}65.75\% &\cellcolor[HTML]{ffb500}66.77\% &\cellcolor[HTML]{ffbb00}67.68\% &\cellcolor[HTML]{ffb900}67.43\% &\cellcolor[HTML]{ffae00}65.65\% \\
\textbf{$\tau_{mem}=420$ $(\beta \approx 0.967)$} &\cellcolor[HTML]{ffea00}75.06\% &\cellcolor[HTML]{ffe600}74.51\% &\cellcolor[HTML]{ffe000}73.58\% &\cellcolor[HTML]{ffd000}71.02\% &\cellcolor[HTML]{ffba00}67.63\% \\
\textbf{$\tau_{mem}=700$ $(\beta \approx 0.980)$} &\cellcolor[HTML]{fff700}77.20\% &\cellcolor[HTML]{ffee00}75.79\% &\cellcolor[HTML]{ffe100}73.79\% &\cellcolor[HTML]{ffd100}71.19\% &\cellcolor[HTML]{ffb500}66.76\% \\
\textbf{$\tau_{mem}=1120$ $(\beta \approx 0.987)$} &\cellcolor[HTML]{fff500}76.88\% &\cellcolor[HTML]{ffed00}75.56\% &\cellcolor[HTML]{ffe400}74.26\% &\cellcolor[HTML]{ffd800}72.35\% &\cellcolor[HTML]{ffb600}67.00\% \\
\textbf{$\tau_{mem}=1680$ $(\beta \approx 0.992)$} &\cellcolor[HTML]{fff300}76.50\% &\cellcolor[HTML]{fff500}76.94\% &\cellcolor[HTML]{ffef00}75.99\% &\cellcolor[HTML]{ffda00}72.61\% &\cellcolor[HTML]{ffba00}67.54\% \\ \hline
\textbf{$\tau_{mem}=\infty$ $(\beta \approx 1)$} &\cellcolor[HTML]{ffff00}78.36\%\tnote{*} &\multicolumn{4}{c}{} \\
\bottomrule
\end{tabular}
\begin{tablenotes}
\item [*] IF Neuron.
\end{tablenotes}
\end{threeparttable}
\end{subtable}
\begin{subtable}{\textwidth}
\centering
\caption{N-MNIST}
\label{sub:mnist-cuba-fsnn}
\begin{threeparttable}
\begin{tabular}{lc|ccccc}\toprule
\textbf{} &\textbf{LIF} &\multicolumn{4}{c}{\textbf{CUBA-LIF}} \\
\textbf{(ms)} &\textbf{$\tau_{syn}=0$ $(\alpha \approx 0)$} &\textbf{$\tau_{syn}=14$ $(\alpha \approx 0.368)$} &\textbf{$\tau_{syn}=28$ $(\alpha \approx 0.606)$} &\textbf{$\tau_{syn}=70$ $(\alpha \approx 0.818)$} &\textbf{$\tau_{syn}=140$ $(\alpha \approx 0.905)$} \\\midrule
\textbf{$\tau_{mem}=14$ $(\beta \approx 0.368)$} &\cellcolor[HTML]{ff1900}96.00\% &\cellcolor[HTML]{ffb800}97.14\% &\cellcolor[HTML]{ffcb00}97.27\% &\cellcolor[HTML]{ffa200}96.98\% &\cellcolor[HTML]{ff9a00}96.92\% \\
\textbf{$\tau_{mem}=70$ $(\beta \approx 0.818)$} &\cellcolor[HTML]{ffb300}97.10\% &\cellcolor[HTML]{ffc200}97.21\% &\cellcolor[HTML]{ffbf00}97.19\% &\cellcolor[HTML]{ff8300}96.76\% &\cellcolor[HTML]{ff5c00}96.48\% \\
\textbf{$\tau_{mem}=140$ $(\beta \approx 0.905)$} &\cellcolor[HTML]{ffe400}97.45\% &\cellcolor[HTML]{ffdb00}97.39\% &\cellcolor[HTML]{ffa900}97.03\% &\cellcolor[HTML]{ff7700}96.67\% &\cellcolor[HTML]{ff4700}96.33\% \\
\textbf{$\tau_{mem}=420$ $(\beta \approx 0.967)$} &\cellcolor[HTML]{fffd00}97.63\% &\cellcolor[HTML]{ffd900}97.37\% &\cellcolor[HTML]{ff9700}96.90\% &\cellcolor[HTML]{ff3c00}96.25\% &\cellcolor[HTML]{ff0900}95.89\% \\
\textbf{$\tau_{mem}=700$ $(\beta \approx 0.980)$} &\cellcolor[HTML]{ffe800}97.48\% &\cellcolor[HTML]{ffde00}97.41\% &\cellcolor[HTML]{ff9e00}96.95\% &\cellcolor[HTML]{ff4000}96.28\% &\cellcolor[HTML]{ff2400}96.08\% \\
\textbf{$\tau_{mem}=1120$ $(\beta \approx 0.987)$} &\cellcolor[HTML]{ffe800}97.48\% &\cellcolor[HTML]{ffd900}97.37\% &\cellcolor[HTML]{ff9c00}96.94\% &\cellcolor[HTML]{ff6200}96.52\% &\cellcolor[HTML]{ff0c00}95.91\% \\
\textbf{$\tau_{mem}=1680$ $(\beta \approx 0.992)$} &\cellcolor[HTML]{ffff00}97.64\% &\cellcolor[HTML]{ffd700}97.36\% &\cellcolor[HTML]{ff9f00}96.96\% &\cellcolor[HTML]{ff4300}96.30\% &\cellcolor[HTML]{ff0000}95.82\% \\ \hline
\textbf{$\tau_{mem}=\infty$ $(\beta \approx 1)$} &\cellcolor[HTML]{ffeb00}97.50\%\tnote{*} &\multicolumn{4}{c}{} \\
\bottomrule
\end{tabular}
\begin{tablenotes}
\item [*] IF Neuron.
\end{tablenotes}
\end{threeparttable}
\end{subtable}
\end{table*}

\subsubsection{Sparsity analysis in \ac{FSNN}}
While learning performance is pivotal, it is also crucial to take into considerations the associated computational cost and energy consumption of using each model which are directly linked to the spiking activity of neurons when using a neuromorphic hardware like Intel Loihi \cite{Davies2018} that computes asynchronously and exploits the sparsity of event-based sensing. To infer an output class, \acp{SNNs} feed the input spikes over a number of time steps and perform event-based synaptic operations only when spike-inputs arrive. These synaptic operations are considered as a metric for benchmarking neuromorphic hardware \cite{Davies2018, Merolla2014}. We explored the impact of the leakages on the sparsity of each model by inferring the test set.  

\begin{figure*}
\centering
\begin{subfigure}{0.98\textwidth}
    \centering
    \includegraphics[width=\textwidth]{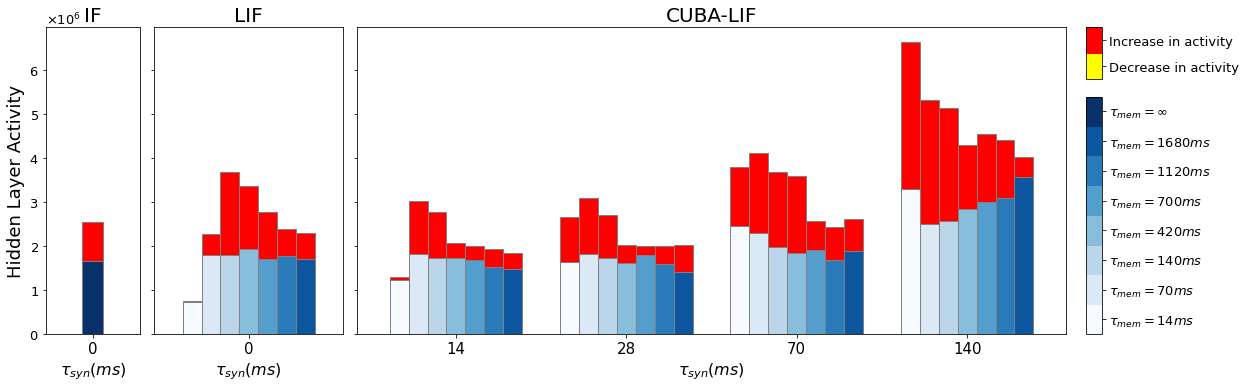}
    \caption{\ac{SHD}}
    \label{fig:shd-spk-rec}
\end{subfigure}
\begin{subfigure}{0.98\textwidth}
    \centering
    \includegraphics[width=\textwidth]{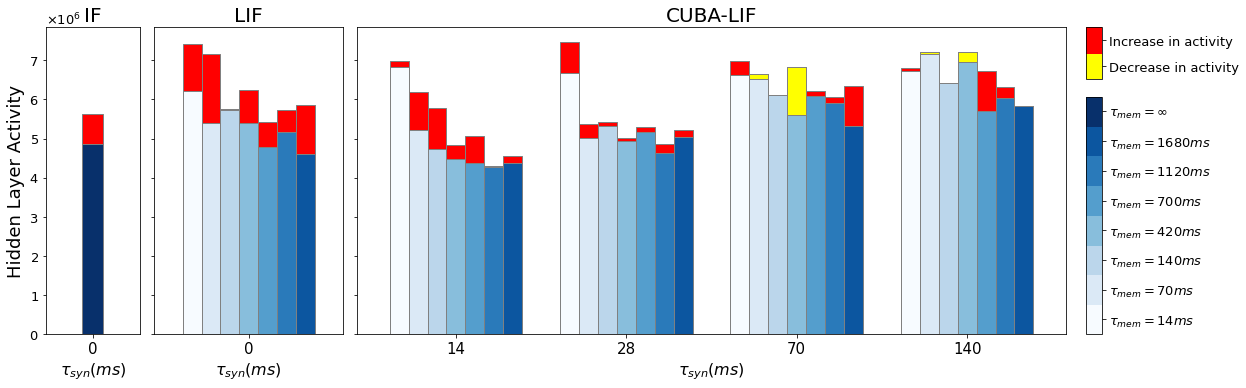}
    \caption{\ac{N-MNIST}}
    \label{fig:nmnist-spk-rec}
\end{subfigure}
\caption{Hidden layer spiking activity with the increase/decrease caused by adding explicit recurrences for each time constants combination of three models. Each grouped set of bars corresponds to one $\tau_{syn}$ value while each bar within the group corresponds to one $\tau_{mem}$ value.}
\label{spk-activity}
\end{figure*}

\ac{SHD} spiking activity recordings in the hidden layer plotted in Figure \ref{spk-activity} show that the time constants combinations that led to the sparsest activity ($\tau_{mem} = 14ms$ for \ac{LIF} and $\tau_{mem}=\tau_{syn}= 14ms$ for \ac{CUBA-LIF}) also resulted in the worst accuracies for both \ac{LIF} and \ac{CUBA-LIF} neurons. This is due to the fast decays in both membrane potential and synaptic current that result in not having enough spikes to hold the information. The same trend however cannot be observed for the \ac{N-MNIST}. This could be due to the fact that leakage parameters do not have a significant impact on spatial information. Nevertheless, we can see an increase in spiking activity with higher values of $\tau_{syn}$ for the \ac{CUBA-LIF} neuron on both datasets. Although it is more apparent on the \ac{SHD}. This increase in spiking activity that resulted in a decrease in accuracy, is associated with \ac{CUBA-LIF} neurons' ability to sustain input spikes over longer durations. However, more spikes do not necessarily lead to better performance, at least for our datasets. These results suggest that there is a sweet-spot where a sufficient number of spikes leads to a an optimal accuracy.

\begin{figure}[]
\centering
\begin{subfigure}{0.5\columnwidth}
    \centering
    \includegraphics[width=\textwidth]{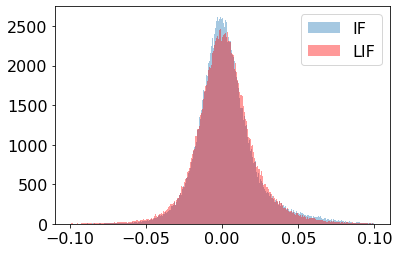}
    \caption{}
\end{subfigure}%
\begin{subfigure}{0.5\columnwidth}
    \centering
    \includegraphics[width=\textwidth]{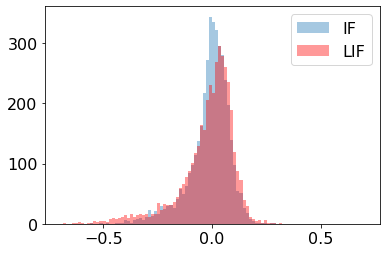}
    \caption{}
\end{subfigure}
\caption{Trained weights distributions for \ac{IF} vs. \ac{LIF} for the weight matrices (a) $W^{(1)}$ and (b) $W^{(2)}$. The standard deviation of $W^{(2)}$ in \ac{LIF} ($12.58\times10^{-2}$) is higher than \ac{IF} ($9.48\times10^{-2}$).}
\label{fig:if-lif-trained-weights}
\end{figure}

Intuitively, we can assume that if all three neurons were to receive the same weighted sum of input, \ac{LIF} neurons would produce comparatively sparser outputs due to their infinite synaptic leak and the layer-wise decay of spikes caused by its membrane leak that acts as a forgetting mechanism. For both datasets, we can see from Figure \ref{spk-activity} that \ac{IF} neurons produced slightly less spikes than \ac{LIF} neurons in some experiments, which is counter-intuitive. In an attempt to understand the cause of this misleading intuition, we plotted the distributions of the trained weights for the \ac{LIF} experiment that has the highest spiking activity ($\tau_{mem} = 420ms$) to compare it with the weights distribution of the \ac{IF} as depicted in Figure \ref{fig:if-lif-trained-weights}. Because it is hard to inspect the distributions visually, we calculated the mean and standard deviation. We can see that the standard deviation of the \ac{LIF}'s weight matrices $W^{(2)}$ is higher than that of the \ac{IF}. This result suggests that \ac{BPTT} tailors the \ac{LIF} model to increase the synaptic weights beyond what is needed for the \ac{IF} model.

\begin{figure}[]
\centering
\begin{subfigure}{0.5\columnwidth}
    \centering
    \includegraphics[width=\textwidth]{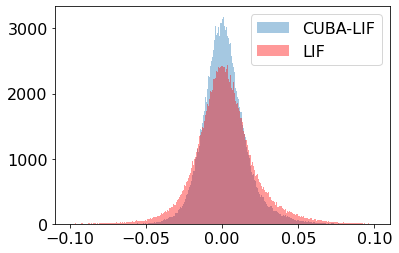}
    \caption{}
\end{subfigure}%
\begin{subfigure}{0.5\columnwidth}
    \centering
    \includegraphics[width=\textwidth]{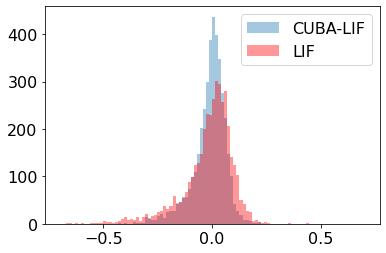}
    \caption{}
\end{subfigure}
\caption{Trained weights distributions for \ac{LIF} vs. \ac{CUBA-LIF} for the weight matrices (a) $W^{(1)}$ and (b) $W^{(2)}$. The standard deviation in \ac{LIF} ($1.99\times10^{-2}$ for $W^{(1)}$ and $1.15\times10^{-1}$ for $W^{(2)}$) is higher than CUBA-LIF ($1.44\times10^{-2}$ for $W^{(1)}$ and $0.77\times10^{-1}$ for $W^{(2)}$).}
\label{fig:CUBA-lif-trained-weights}
\end{figure}

The \ac{CUBA-LIF} model, on the other hand, was able achieve the sparsest activity among the three models for certain combinations of time constants despite its ability to sustain input spikes for longer duration. To that effect, we plotted the weights distributions of \ac{CUBA-LIF}'s experiment with $\tau_{mem} = 1680ms$ and $\tau_{syn} = 28ms$ and compared it with that of the \ac{LIF} experiment that has the same $\tau_{mem}$ value. Again, we calculated the mean and standard deviation. We can see that the standard deviation of \ac{CUBA-LIF}'S weight matrices $W^{(2)}$ is higher than that of the \ac{LIF}. Once more, this can be attributed to \ac{BPTT}. Therefore, the results clearly indicate that the leakages do not necessarily lead to sparser activity.

\subsection{Impact of explicit recurrences}
To study the effect of recurrences on learning spatio-temporal patterns, we added explicit recurrent connections to neurons in the hidden layer and confronted the three neuron models in the context of a \ac{RSNN}. Similar to the experiments in the \ac{FSNN}, we performed a grid search across the same combinations of time constants for \ac{CUBA-LIF}, a sweep for the same $\tau_{mem}$ values for \ac{LIF}, and the same experiments for \ac{IF}.

\subsubsection{Accuracy analysis in \ac{RSNN}}
\begin{figure}
\centering
\begin{subfigure}{0.5\columnwidth}
    \centering
    \includegraphics[width=\textwidth]{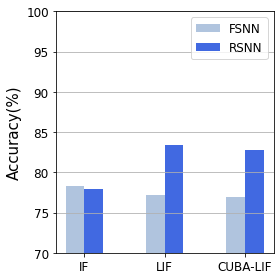}
    \caption{\ac{SHD}}
    \label{fig:best-compare-nmnist}
\end{subfigure}%
\begin{subfigure}{0.5\columnwidth}
    \centering
    \includegraphics[width=\textwidth]{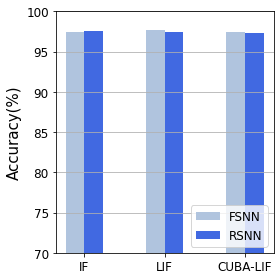}
    \caption{\ac{N-MNIST}}
    \label{fig:best-compare-shd}
\end{subfigure}
\caption{Best accuracies comparison between models in \ac{FSNN} vs. \ac{RSNN}.}
\label{best-compare}
\end{figure}

As shown in Table \ref{tab:cuba-rsnn}, results of the \ac{SHD} dataset show that recurrent architectures reached a significantly higher performances than their feed-forward counterparts across all combinations of time constants for both \ac{CUBA-LIF} and \ac{LIF}. However, that is not the case for the \ac{N-MNIST}. This is not surprising knowing the inherent ability of \ac{RCNN} to handle time series and sequential data. For both datasets however, we can still observe the same trend as in \ac{FSNN} such that  $\tau_{mem}$ values below $420ms$ result in a significant decrease in accuracy for both \ac{CUBA-LIF} and \ac{LIF}, while \ac{CUBA-LIF} performed better with smaller values of $\tau_{syn}$. Nevertheless, the \ac{LIF} neuron reached the highest accuracy among the two. For the \ac{IF} neuron, adding explicit recurrences reduced the accuracy by 0.43\% on \ac{SHD} and lead to comparable accuracy on \ac{N-MNIST}. A comparison between the best accuracies obtained by the models in both \ac{FSNN} and \ac{RSNN} is presented if Figure \ref{best-compare}. 

Given \ac{IF}'s good performance in \ac{FSNN} with both datasets and inferior performance in \ac{RSNN} with \ac{SHD}, it becomes clear that leakages are important when there are both temporal information in the data and a recurrent topology in the network. This result is the most important finding of this work and our unique contribution to the neuromorphic computing literature. To the best of our knowledge, the highest accuracies we were able to reach on the \ac{SHD} dataset are very close to state-of-the-art results \cite{Dampfhoffer2022}. Table \ref{tab:literature-compare} compares our best results with other works in the literature.

\begin{table*}[!htp]\centering
\caption{Three neuron models accuracy in RSNN.}\label{tab:cuba-rsnn}
\scriptsize
\begin{subtable}{\textwidth}
\centering
\caption{SHD}
\begin{threeparttable}
\begin{tabular}{lc|ccccc}\toprule
\textbf{} &\textbf{LIF} &\multicolumn{4}{c}{\textbf{CUBA-LIF}} \\
\textbf{(ms)} &\textbf{$\tau_{syn}=0$ $(\alpha \approx 0)$} &\textbf{$\tau_{syn}=14$ $(\alpha \approx 0.368)$} &\textbf{$\tau_{syn}=28$ $(\alpha \approx 0.606)$} &\textbf{$\tau_{syn}=70$ $(\alpha \approx 0.818)$} &\textbf{$\tau_{syn}=140$ $(\alpha \approx 0.905)$} \\\midrule
\textbf{$\tau_{mem}=14$ $(\beta \approx 0.368)$} &\cellcolor[HTML]{ff2400}44.67\% &\cellcolor[HTML]{ff7200}58.56\% &\cellcolor[HTML]{ffc700}73.54\% &\cellcolor[HTML]{ffd000}75.26\% &\cellcolor[HTML]{ffc500}73.19\% \\
\textbf{$\tau_{mem}=70$ $(\beta \approx 0.818)$} &\cellcolor[HTML]{ffb600}70.51\% &\cellcolor[HTML]{ffd700}76.41\% &\cellcolor[HTML]{ffe900}79.64\% &\cellcolor[HTML]{ffe800}79.41\% &\cellcolor[HTML]{ffcd00}74.59\% \\
\textbf{$\tau_{mem}=140$ $(\beta \approx 0.905)$} &\cellcolor[HTML]{ffe200}78.34\% &\cellcolor[HTML]{ffee00}80.48\% &\cellcolor[HTML]{fff200}81.25\% &\cellcolor[HTML]{ffe400}78.64\% &\cellcolor[HTML]{ffd400}75.96\% \\
\textbf{$\tau_{mem}=420$ $(\beta \approx 0.967)$} &\cellcolor[HTML]{fffb00}82.72\% &\cellcolor[HTML]{fff600}81.96\% &\cellcolor[HTML]{fff500}81.71\% &\cellcolor[HTML]{ffdb00}77.05\% &\cellcolor[HTML]{ffd000}75.15\% \\
\textbf{$\tau_{mem}=700$ $(\beta \approx 0.980)$} &\cellcolor[HTML]{fffd00}83.06\% &\cellcolor[HTML]{fff900}82.44\% &\cellcolor[HTML]{ffef00}80.65\% &\cellcolor[HTML]{ffe500}78.81\% &\cellcolor[HTML]{ffd400}75.91\% \\
\textbf{$\tau_{mem}=1120$ $(\beta \approx 0.987)$} &\cellcolor[HTML]{fffe00}83.24\% &\cellcolor[HTML]{fffb00}82.74\% &\cellcolor[HTML]{ffef00}80.73\% &\cellcolor[HTML]{ffe500}78.89\% &\cellcolor[HTML]{ffd200}75.52\% \\
\textbf{$\tau_{mem}=1680$ $(\beta \approx 0.992)$} &\cellcolor[HTML]{ffff00}83.41\% &\cellcolor[HTML]{fff800}82.25\% &\cellcolor[HTML]{ffef00}80.68\% &\cellcolor[HTML]{ffea00}79.83\% &\cellcolor[HTML]{ffd500}76.06\% \\ \hline
\textbf{$\tau_{mem}=\infty$ $(\beta \approx 1)$} &\cellcolor[HTML]{ffe000}77.93\%\tnote{*} &\multicolumn{4}{c}{} \\
\bottomrule
\end{tabular}
\begin{tablenotes}
\item [*] IF Neuron.
\end{tablenotes}
\end{threeparttable}
\end{subtable}
\begin{subtable}{\textwidth}
\centering
\caption{N-MNIST}
\begin{threeparttable}
\begin{tabular}{lc|ccccc}\toprule
&\textbf{LIF} &\multicolumn{4}{c}{\textbf{CUBA-LIF}} \\
\textbf{(ms)} &\textbf{$\tau_{syn}=0$ $(\alpha \approx 0)$} &\textbf{$\tau_{syn}=14$ $(\alpha \approx 0.368)$} &\textbf{$\tau_{syn}=28$ $(\alpha \approx 0.606)$} &\textbf{$\tau_{syn}=70$ $(\alpha \approx 0.818)$} &\textbf{$\tau_{syn}=140$ $(\alpha \approx 0.905)$} \\\midrule
\textbf{$\tau_{mem}=14$ $(\beta \approx 0.368)$} &\cellcolor[HTML]{ff4e00}96.18\% &\cellcolor[HTML]{ffd500}97.22\% &\cellcolor[HTML]{ffcb00}97.14\% &\cellcolor[HTML]{ffa300}96.83\% &\cellcolor[HTML]{ffae00}96.92\% \\
\textbf{$\tau_{mem}=70$ $(\beta \approx 0.818)$} &\cellcolor[HTML]{ffc600}97.10\% &\cellcolor[HTML]{ffdc00}97.27\% &\cellcolor[HTML]{ffc400}97.09\% &\cellcolor[HTML]{ff9c00}96.78\% &\cellcolor[HTML]{ff5d00}96.29\% \\
\textbf{$\tau_{mem}=140$ $(\beta \approx 0.905)$} &\cellcolor[HTML]{ffdd00}97.28\% &\cellcolor[HTML]{ffd800}97.24\% &\cellcolor[HTML]{ffb900}97.00\% &\cellcolor[HTML]{ff8700}96.62\% &\cellcolor[HTML]{ff4500}96.11\% \\
\textbf{$\tau_{mem}=420$ $(\beta \approx 0.967)$} &\cellcolor[HTML]{ffeb00}97.39\% &\cellcolor[HTML]{ffda00}97.26\% &\cellcolor[HTML]{ffd000}97.18\% &\cellcolor[HTML]{ff5a00}96.27\% &\cellcolor[HTML]{ff0000}95.57\% \\
\textbf{$\tau_{mem}=700$ $(\beta \approx 0.980)$} &\cellcolor[HTML]{fff200}97.44\% &\cellcolor[HTML]{ffe600}97.35\% &\cellcolor[HTML]{ffa000}96.81\% &\cellcolor[HTML]{ff4200}96.08\% &\cellcolor[HTML]{ff2800}95.88\% \\
\textbf{$\tau_{mem}=1120$ $(\beta \approx 0.987)$} &\cellcolor[HTML]{fff700}97.48\% &\cellcolor[HTML]{ffe200}97.32\% &\cellcolor[HTML]{ff9700}96.74\% &\cellcolor[HTML]{ff5100}96.20\% &\cellcolor[HTML]{ff1300}95.72\% \\
\textbf{$\tau_{mem}=1680$ $(\beta \approx 0.992)$} &\cellcolor[HTML]{ffee00}97.41\% &\cellcolor[HTML]{ffd500}97.22\% &\cellcolor[HTML]{ff9f00}96.80\% &\cellcolor[HTML]{ff5200}96.21\% &\cellcolor[HTML]{ff3100}95.95\% \\ \hline
\textbf{$\tau_{mem}=\infty$ $(\beta \approx 1)$} &\cellcolor[HTML]{ffff00}97.54\%\tnote{*} &\multicolumn{4}{c}{} \\
\bottomrule
\end{tabular}
\begin{tablenotes}
\item [*] IF Neuron.
\end{tablenotes}
\end{threeparttable}
\end{subtable}
\end{table*}

\subsubsection{Sparsity analysis in \ac{RSNN}}
Similar to what we did in \ac{FSNN}, we recorded the spiking activity of neurons in the hidden layer when the test set is inferred. \ac{SHD} spikes count recordings plotted in Figure \ref{fig:shd-spk-rec} show that explicit recurrent connections increase activity in all neurons for every combination of time constants. On average, we have 53.55\% increase in spiking activity for \ac{CUBA-LIF}, 53.35\% for \ac{LIF}, and 53.58\% for \ac{IF}. \ac{N-MNIST} spikes count, on the other hand, did not increase for every combination time constants. It even decreased for some as shown in Figure \ref{fig:nmnist-spk-rec}. On average, we have 3.89\% increase for \ac{CUBA-LIF}, 16.78\% for \ac{LIF}, and 15.75\% for \ac{IF}.

It is hard to say whether or not the bigger increase in spiking activity for \ac{SHD} contributed to its improved classification accuracy given that we saw similar increase for \ac{IF} neurons but a worsened performance. Given the \ac{CUBA-LIF} experiments that resulted in classification performance that is almost as good as that of the \ac{LIF} also resulted in the slightest increase in spiking activity. The \ac{CUBA-LIF} model could be more suitable for low power applications especially if tuned better to reach even higher accuracy. 

\begin{table}
\centering
\caption{Our best \ac{SHD} results compared to \cite{Cramer2022}, \cite{Perez-Nieves_2021}, and \cite{Dampfhoffer2022}}
\label{tab:literature-compare}
\begin{tabular}{lccc} 
\toprule
& \textbf{Neuron} & \textbf{Standard} & \textbf{Heterogeneous}  \\ 
& \textbf{model} & \textbf{training} & \textbf{training}  \\ 
\midrule
\textbf{\cite{Cramer2022}}                         & \ac{CUBA-LIF}              & 79.9\%                     & -                                \\ 
\hline
\textbf{\cite{Perez-Nieves_2021}}                         & \ac{CUBA-LIF}              & 71.7\%                     & 82.7\%                           \\ 
\hline
\multirow{2}{*}{\textbf{\cite{Dampfhoffer2022}}}        & \ac{CUBA-LIF}              & \textbf{83.7\%}            & -                                \\ \cline{2-4}
                                    & \ac{LIF}                   & 80.6\%                     & -                                \\ 
\hline
\multirow{2}{*}{\textbf{This work}} & \ac{CUBA-LIF}              & 82.74\%                    & 82.84\%                          \\ \cline{2-4}
                                    & \ac{LIF}                   & \textbf{83.41\%}           & \textbf{83.47\%}                 \\
\bottomrule
\end{tabular}
\end{table}

\subsection{Impact of neural heterogeneity}
Most existing learning methods learn the synaptic weights only while requiring a manual tuning of leakages-related parameters similar to our previously presented experiments. These parameters are chosen to be the same for all neurons, which could limit the diversity and expressiveness of \acp{SNNs}. 
In biological brains, neuronal cells have different time constants with distinct stereotyped distributions depending on the cell type \cite{Manis2019, Manis_Kasten2019, Hawrylycz2012}.
To assess whether this heterogeneity plays an important functional role or is just a byproduct of noisy developmental processes, we incorporated learnable time constants in the training process. Hence, $\tau_{mem}$ and $\tau_{syn}$ will not be treated as hyper-parameters, but learned parameters along with the synaptic weights. We refer to this training process as heterogeneous training. Since the \ac{IF} neuron has fixed values of time constants: $\tau_{mem} = \infty$ and $\tau_{syn} = 0$, it is not concerned with heterogeneous training. On the other hand, \ac{LIF} neuron has a fixed $\tau_{syn}$ equal to zero but a variable $\tau_{mem}$ which we were able to train. For \ac{CUBA-LIF}, both time constants are trained.

To evaluate the performance of incorporating learnable time constants in comparison with the standard training in our previously presented experiments, we initialized $\tau_{mem}$ and $\tau_{syn}$ to the same values we used in our grid search for both \ac{CUBA-LIF} and \ac{LIF} and trained then along with the synaptic weights. We also conducted these experiments in both \ac{FSNN} and \ac{RSNN}. We found that incorporating learnable time constants did not have a profound impact on both datasets. As can be seen in Table \ref{tab:compare-standard-hetero}, the best accuracies obtained with heterogeneous training are slightly higher than that of the standard training for the \ac{SHD}. Conversely, \ac{N-MNIST} reached the best accuracies with standard training. This result tells us that heterogeneity in time constants could further improve performance for data with information content in their temporal dynamics.

\begin{table}
\centering
\caption{Comparison between best accuracies of standard vs. heterogeneous training.}
\begin{subtable}{\columnwidth}
\centering
\caption{\ac{FSNN}}
\begin{tabular}{lccc} 
\toprule
& \textbf{Neuron} & \textbf{Standard} & \textbf{Heterogeneous}  \\ 
& \textbf{model} & \textbf{training} & \textbf{training}  \\ 
\midrule
\multirow{2}{*}{\textbf{\ac{SHD}}}     & \ac{CUBA-LIF}              & 76.94\%                    & \textbf{78.69\%}                 \\ 
\cline{2-4}
                                  & \ac{LIF}                   & 77.20\%                    & \textbf{79.84\%}                 \\ 
\midrule
\multirow{2}{*}{\textbf{\ac{N-MNIST}}} & \ac{CUBA-LIF}              & \textbf{97.41\%}           & 97.24\%                          \\ 
\cline{2-4}
                                  & \ac{LIF}                   & \textbf{97.64\%}           & 97.41\%                          \\
\bottomrule
\end{tabular}
\end{subtable}
\begin{subtable}{\columnwidth}
\centering
\caption{\ac{RSNN}}
\begin{tabular}{lccc} 
\toprule
& \textbf{Neuron} & \textbf{Standard} & \textbf{Heterogeneous}  \\ 
& \textbf{model} & \textbf{training} & \textbf{training}  \\ 
\midrule
\multirow{2}{*}{\textbf{\ac{SHD}}}     & \ac{CUBA-LIF}              & 82.74\%                    & \textbf{82.84\%}                 \\ 
\cline{2-4}
                                  & \ac{LIF}                   & 83.41\%                    & \textbf{83.47\%}                 \\ 
\midrule
\multirow{2}{*}{\textbf{\ac{N-MNIST}}} & \ac{CUBA-LIF}              & \textbf{97.35\%}           & 97.14\%                          \\ 
\cline{2-4}
                                  & \ac{LIF}                   & \textbf{97.48\%}           & 97.38\%                          \\
\bottomrule
\end{tabular}
\end{subtable}
\label{tab:compare-standard-hetero}
\end{table}

\begin{table}
\centering
\caption{Number of multiplication, addition and comparison operations per spiking neuron at each time step, where N is the number of inputs (feedforward and/or recurrent) to the neuron and P is the percentage of those inputs that receive a spike.}
\label{tab:neurons_complexity}
\begin{tabular}{lccc}
\toprule
\textbf{Neuron model} & \textbf{IF} & \textbf{LIF} & \textbf{CUBA-LIF} \\
\hline
Multiplications       & 0           & 1            & 2                 \\
\hline
Additions             & $N\times P$   & $N\times P$    & $N\times P + 1$     \\
\hline
Comparisons           & 1           & 1            & 1                 \\   
\bottomrule
\end{tabular}
\end{table}


\section{Discussion}
In the neuro-scientific literature, it has been reported that leakages in biological neurons exist in many contexts such as synaptic transmission in the visual cortex \cite{Artun1998} and sodium ion channels \cite{Ren_2011, Snutch_Monteil2011}. Many spiking neuron models imitate this leaky behaviour through an exponential decay in the synaptic current and membrane potential. Other models prioritize computational efficiency by removing the leakage. To tackle the lack in understanding of the effect of these leakages from the modeling perspective, we confronted three spiking neuron models with variable degrees of leaky behaviour, namely the \ac{CUBA-LIF}, \ac{LIF}, and \ac{IF}, in classification tasks with a number of degrees of freedom.

We first trained \acp{SNNs} using the three neuron models with a feed-forward network to classify visual patterns of written digits from the \ac{N-MNIST} dataset and auditory information of spoken digits from the \ac{SHD} datasets. Surprisingly, the \ac{IF} model, despite the absence of leaky behaviour and the resulting lack of inherent temporal dynamics, slightly outperformed the other models on the \ac{SHD} by reaching an accuracy of 78.36\% $\pm$ 0.87\%, and closely matched the best of \ac{LIF} model accuracy on the \ac{N-MNIST} by reaching 97.50\% $\pm$ 0.06\%. \ac{CUBA-LIF} on the other hand, had the inferior performance among the three models on both datasets despite its intrinsic temporal dynamics caused by both synaptic and membrane leaks. Both \ac{LIF} and \ac{CUBA-LIF} saw a drastic decrease in accuracy when $\tau_{mem}$ is less than $420ms$, which leads to a fast decay in membrane potential and loss of information. We also found that \ac{CUBA-LIF} reached its highest accuracies when its dynamics are close to those of the \ac{LIF}. We conclude that leakages do not necessarily lead to improved performances even on temporally complex tasks when using feed-forward networks. 
In terms of sparsity, it is \ac{IF} to see sparser activity in \ac{IF} neurons and \ac{CUBA-LIF} neurons with smaller values of $\tau_{syn}$ than their \ac{LIF} counterpart. Upon inspection of the trained weights distributions, it seems that \ac{BPTT} is tailoring \ac{LIF} neurons to have bigger weights, and hence more spikes.
Therefore, leakages do not always lead to sparser activity. Furthermore, we noticed that very low spiking activity resulted in the worst classification performance on the \ac{SHD}. Very high spiking activity associated with bigger $\tau_{syn}$ values also resulted in a worsened performance. These results suggest that there is a sweet-spot where a sufficient amount of spikes produce an optimal classification accuracy.

Overall, \ac{IF} neurons are sufficient when using data without temporal information or a network without recurrence in terms of classification accuracy and sparsity. It suggests that the fundamental ingredient of spiking neurons is their statefullness, i.e. having an internal state with an implicit recurrence, even without leakage.
Furthermore, they offer a better alternative if we consider digital neuromorphic hardware design that is based on application-specific needs. \ac{IF} neurons could be very cheap in terms of hardware resources, as they only perform additions for the input integration and a comparison for the output evaluation. In contrast, the \ac{LIF} and \ac{CUBA-LIF} neurons require multipliers to implement the leakage in their current and/or voltage compartments as shown in Table \ref{tab:neurons_complexity}, thus resulting in more expensive hardware.
Next, we added explicit recurrent connections to the neurons in the hidden layer. Expectedly, we saw a big improvement in accuracy for the \ac{SHD} that has a rich temporal structure and no improvement at all for the \ac{N-MNIST} that hos mostly spatial structure. However, recurrences did not have any impact on the \ac{IF} neuron on both datasets. Therefore, we conclude that the inherent temporal dynamics introduced by the leakages are only necessary when we use both data with a rich temporal structure and a neural network with a explicit recurrence. The best \ac{SHD} accuracies we were able to obtained in a \ac{RSNN} were very close to state-of-the-art results \cite{Dampfhoffer2022} such that we reached 82.74\% $\pm$ 0.17\% with \ac{CUBA-LIF} and 83.41\% $\pm$ 0.37\% with the \ac{LIF}.
In terms of sparsity, we saw a bigger increase in spiking activity with the \ac{SHD} than the \ac{N-MNIST}. In both datasets, the \ac{CUBA-LIF} neurons with the best time constants combinations added the smallest number of spikes, which gives them an advantage in sparsity compared to \ac{LIF} neurons.

Finally, we introduced heterogeneity in the considered spiking neurons by incorporating learnable time constants in the training process along with the synaptic weights. This heterogeneous training slightly improved performance on the \ac{SHD}, which has a complex temporal structure. The best \ac{SHD} accuracies we obtained with heterogeneous training in \ac{RSNN} were also very close to state-of-the-art results \cite{Dampfhoffer2022} with 82.84\% $\pm$ 1.17\% for \ac{CUBA-LIF} and 83.47\% $\pm$ 2.12\% for \ac{LIF}.


\section{Conclusion}
In this work we explored the effect of spiking neurons synaptic and membrane leakages, network explicit recurrences and time constants heterogeneity on event-based spatio-temporal pattern recognition. The main findings of our work can be summarized as follows: 

\begin{itemize}
    \item Neural leakages are only important when there are both temporal information in the data and explicit recurrent connections in the network.
    \item Neural leakages do not necessarily lead to sparser spiking activity in the network.
    \item Time constants heterogeneity slightly improves performance on data with a rich temporal structure and does not affect performance on data with a spatial structure.
\end{itemize}

This work supports the identification of the right level of model abstraction of biological evidences needed to build efficient application-specific neuromorphic hardware. This is a crucial analysis for advancing the field beyond state-of-the-art, especially when constrains on resources are critical (e.g. edge computing). In fact, when using digital neuromorphic architectures, \ac{IF} neurons have been shown to be 2 $\times$ smaller and more power-efficient than formal Perceptrons \cite{Khacef_Abderrahmane2018}. It is nevertheless not clear how this gain evolves when adding a multiplier to implement a \ac{LIF} or \ac{CUBA-LIF} neuron. Further works will focus on implementing these two architectures in FPGAs for fast prototyping.
In addition, \ac{IF} neurons give the possibility to implement a digital asynchronous processing purely driven by the input, since there is no inherent temporal dynamics in the spiking neurons. On the other hand, \ac{LIF} and \ac{CUBA-LIF} neurons require algorithmic time-steps where the leakage is updated regardless of the presence of input spikes. Further works will explore the impact of both paradigms in energy-efficiency on the Loihi neuromorphic chip \cite{Davies2018}.

Furthermore, it is important to mention that our results only hold in benchmarking so far. In a real-world scenario such as continuous keyword spotting, there can be more noise in the data but also in void. Hence, when using the \ac{IF} neurons that do not have any leakage, this noise can accumulate and create false positives and degrade the performance. Indeed, the low-pass filtering effect of the spiking neurons leakages has been shown to eliminate high frequency components from the input and enhance the noise robustness of \acp{SNNs}, especially in real-world environments \cite{Chowdhury_Lee2021}.
In addition, given that the \ac{LIF} model achieved a superior performance when compared to the \ac{CUBA-LIF}, it is important to investigate where the latter could perform better. More complex tasks could show such a gain for the \ac{CUBA-LIF} neuron, because of its current compartment which is an extra state that gives more potential for spatio-temporal feature extraction.
Finally, spiking neural networks in neuromorphic hardware can be used beyond fast and efficient inference, by adding adaptation through local synaptic plasticity \cite{Qiao_etal15,Quintana_etal22,Khacef_etal22}. In this context, the impact of the leakage can be different, as the inherent temporal dynamics is required in some plasticity mechanisms \cite{Brader_etal07,Clopath_etal10} for online learning.


\section*{Acknowledgements}
We would like to thank Dylan Muir for the fruitful discussion on DynapCNN.
We would like to thank the Institute of Electrical and Electronic Engineering at the University of Boumerdes for supporting this work.
We would also like to acknowledge the financial support of the CogniGron research center and the Ubbo Emmius Funds of the University of Groningen.


\bibliographystyle{unsrt}
\bibliography{biblio}

\end{document}